\documentclass[10pt,twocolumn,letterpaper]{article}

\usepackage{iccv}
\usepackage{times}
\usepackage{epsfig}
\usepackage{graphicx}
\usepackage{amsmath}
\usepackage{amssymb}

\usepackage{booktabs}

\usepackage{pifont}

\usepackage{times}
\usepackage{epsfig}
\usepackage{url}
\usepackage[dvipsnames]{xcolor}
\usepackage{multirow}
\usepackage{wrapfig}
\usepackage{graphicx}
\usepackage{subcaption}
\usepackage{booktabs} %
\newcommand{\ra}[1]{\renewcommand{\arraystretch}{#1}} %
\usepackage{multirow} %
\usepackage{url} %
\usepackage{cuted} %
\usepackage{capt-of}
\usepackage[font={small}]{caption}
\usepackage{epigraph}
\usepackage{enumitem}
\usepackage{xspace}
\usepackage{censor}

\newcommand{\catnet}{CA$^2$T-Net }

\usepackage[pagebackref=true,breaklinks=true,letterpaper=true,colorlinks,bookmarks=false]{hyperref}

\usepackage[capitalize]{cleveref}
\crefname{section}{Sec.}{Secs.}
\Crefname{section}{Section}{Sections}
\Crefname{table}{Table}{Tables}
\crefname{table}{Tab.}{Tabs.}

\iccvfinalcopy %

\def\iccvPaperID{8599} %

\ificcvfinal\fi

\begin{document}

\title{CA$^2$T-Net: Category-Agnostic 3D Articulation Transfer from Single Image}

\author{
Jasmine Collins\textsuperscript{1} \quad  Anqi Liang\textsuperscript{2} \quad Jitendra Malik\textsuperscript{1} \quad Hao Zhang\textsuperscript{2} \quad Fr\'ed\'eric Devernay\textsuperscript{2} \\
\textsuperscript{1} UC Berkeley \qquad \textsuperscript{2} Amazon
}
\maketitle

\maketitle
\ificcvfinal\thispagestyle{empty}\fi

\begin{abstract}
\vspace{-0.1cm}
We present a neural network approach to transfer the motion from a single image of an articulated object to a {\em rest-state\/} (i.e., unarticulated) 3D model. Our network learns to predict the object's pose, part segmentation, and corresponding motion parameters to reproduce the articulation shown in the input image. The network is composed of three distinct branches that take a shared joint image-shape embedding and is trained end-to-end. Unlike previous methods, our approach is independent of the topology of the object and can work with objects from arbitrary categories. Our method, trained with only synthetic data, can be used to automatically animate a mesh, infer motion from real images, and transfer articulation to functionally similar but geometrically distinct 3D models at test time. 
\vspace{-5mm}
\end{abstract}

\section{Introduction}
\label{sec:intro}

We live in a 3D world where interacting with objects in our environment is necessary to carry out daily activities. Therefore, the objects around us have been designed with the appropriate structures and part motions to afford various actions to both humans and robotic agents. Whether for general visual perception tasks such as functionality inference or robotic action planning, it is a valuable skill to be able to predict the part motions or articulations of everyday objects. Recently, there has 
been much interest in training deep neural networks to learn such predictions, especially in the context of embodied AI~\cite{xiang2020sapien, mo2021where2act}.
However, the availability of 3D models with associated part
articulations~\cite{wang2019shape2motion,xiang2020sapien} for training these methods is still limited.

\begin{figure}[t]
    \centering
    \includegraphics[width=1.0\linewidth]{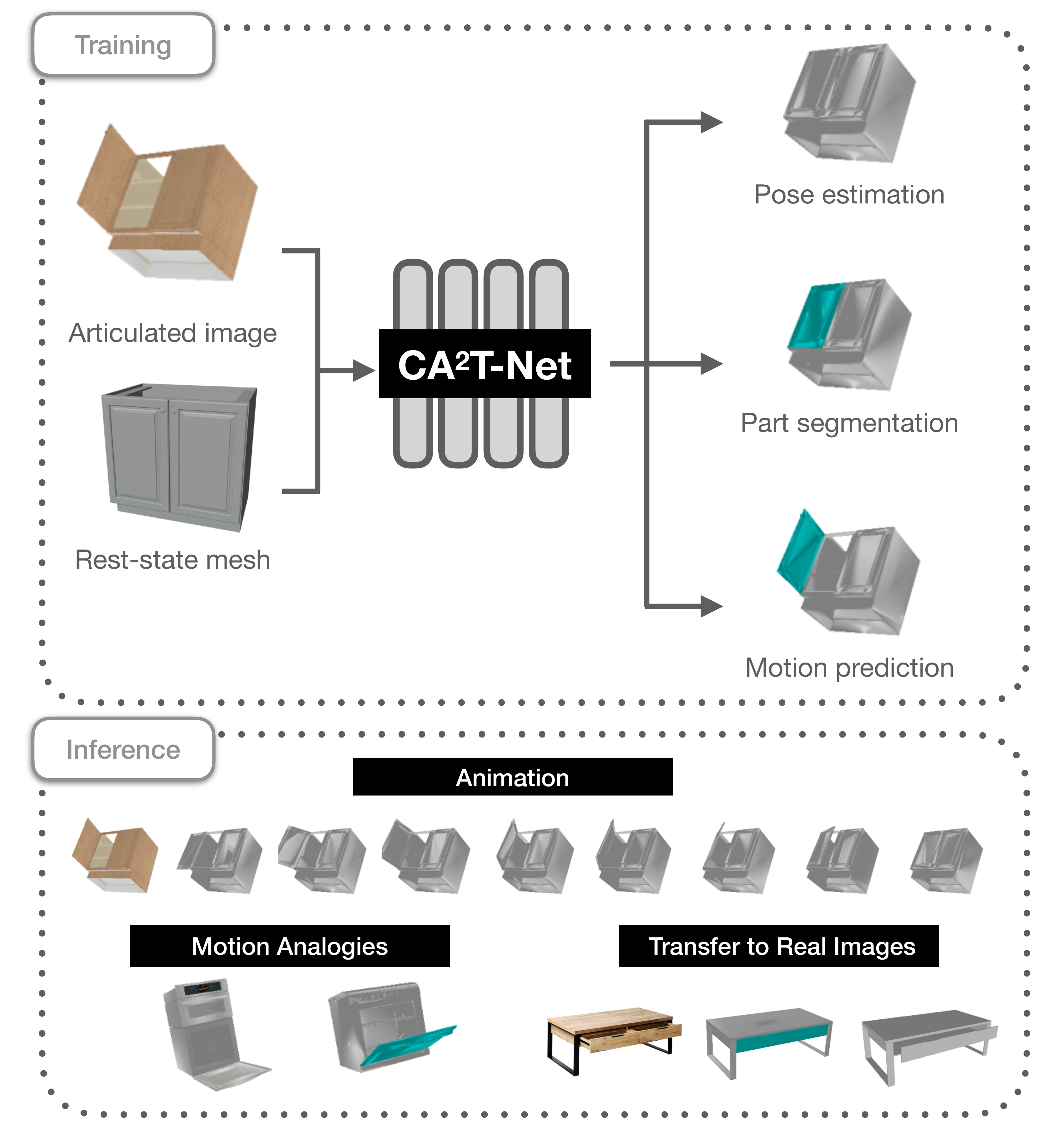}
    \caption{\textbf{Predicting pose, part, and motion annotations for synthetic meshes given an exemplar image.} At test time, we can perform 3D animation, transfer motion to functionally similar objects, and generalize to real \{image, mesh\} pairs.}
    \vspace{-8pt}
    \label{fig:teaser}
\end{figure}

Creating articulated 3D models relies on human effort which can be expensive in terms of time, cost and expertise. As a result, most 3D model datasets do not contain objects with any kind of functional part and motion annotations. For example, Amazon Berkeley Objects (ABO)~\cite{collins2022abo} is a recently published, large-scale
(8K models) 
dataset of 
artist-created 3D models 
of real household objects. While most of the 3D models in ABO represent objects that can be interacted with, such as desks with openable drawers or chairs that recline, the 3D models have all been constructed in their ``rest" (i.e., unarticulated) poses that do not expose their real functionality. The same is true for the vast majority of 3D models in well-established repositories such as ShapeNet~\cite{chang2015shapenet} (3M models) and PartNet~\cite{mo2019partnet} (27K models)
The largest 3D datasets with part articulations, SAPIEN~\cite{xiang2020sapien} and Shape2Motion~\cite{wang2019shape2motion}, only contain 2K 
{\em manually\/} annotated synthetic models.

\begin{table*}[t]
\begin{center}
{
    \setlength\tabcolsep{3pt}
    \ra{1.1} %
\begin{tabular}{llccccc}
  \toprule
  Method & Task & Input & Pose & 3D Part Seg. & Category-Specific \\ 
  \midrule
  DeepPartInduction~\cite{yi2018deep} & Transfer articulation from PC $\rightarrow$ PC & 2 PCs & $\times$ &  $\checkmark$ & No \\ 
  Shape2Motion~\cite{wang2019shape2motion} & Predict motion for all moveable parts & PC &  $\times$ & $\checkmark$ & No \\ 
  RPM-Net~\cite{yan2019rpm} & Predict motion for all moveable parts & PC & $\times$ &  $\checkmark$ & No \\
  ANCSH~\cite{li2020category} & Articulated pose estimation & PC & $\checkmark$ & $\checkmark$ & Yes\\
  NASAM~\cite{wei2022self} & Reconstruct and animate objects & RGBs + pose & $\times$ & $\checkmark$ & Yes \\
  OPD~\cite{jiang2022opd} & Predict motion for all openable parts & RGB & $\checkmark$ & $\times$ & No \\
  \midrule
  \catnet (ours) & Transfer articulation from image $\rightarrow$ mesh & RGB, PC & $\checkmark$ & $\checkmark$ & No \\ 
  \bottomrule
\end{tabular}
}
\end{center}
\vspace*{-5mm}
\caption{\textbf{\catnet addresses the problem of \textit{single-view articulation transfer}, from an image to a mesh}. Existing work targets related problems, but differs in terms of task and inputs of the model.}
\vspace{-5pt}
\label{tab:related-work}
\end{table*}

In this paper, we introduce {\em single-view 3D articulation transfer\/}, a learning-based approach aimed at endowing the many existing 3D models, in particular, those of real products such as ABO, with part articulations. Specifically, given a single RGB image, $I$, showing how an object can be articulated, and a 3D mesh model $M$ in its rest state, our goal is to infer the pose, motion parameters, and 3D part segmentation of $M$ such that we can re-pose and transform it to match the articulation in the input image. Our method is structure-agnostic, as it makes no assumption on the topology of the input model.
Unlike many existing methods for articulated pose estimation, our approach is also \textit{category-agnostic},
meaning no further training is necessary to run our model on objects from novel categories and it can make predictions for objects from arbitrary categories. Further, the object represented in the input image $I$ and the mesh $M$ need not be exactly the same, but should simply share a functional similarity to allow a plausible articulation transfer. To our knowledge, our work is the first to pose the problem of single-view articulation transfer onto a target 3D object. As indicated in Table~\ref{tab:related-work}, the most relevant works~\cite{yi2018deep,wang2019shape2motion,li2020category,jiang2022opd,yan2019rpm} differ from our problem setting in one way or another.

We call our network \catnet for category-agnostic articulation transfer, and demonstrate that our method has superior performance in single-view 3D articulation transfer compared to baseline methods and the current state-of-the-art in terms of motion prediction. Our approach takes as input an articulated image and rest-state 3D point cloud and uses three branches to predict global pose, part segmentation, and 3D motion parameters. The network is trained end-to-end using synthetic data. We also show three downstream applications of our trained approach: automatic mesh animation, generalization to functionally similar objects, and transfer to real objects and images in ABO.

Our work is a first step towards replacing manual part and motion annotations~\cite{wang2019shape2motion,xiang2020sapien} with a fully automated inference of part articulations and motion parameters of arbitrary 3D shapes. With the wide availability of images depicting object articulations (e.g., catalog images found in the product metadata within ABO), our image-to-3D transfer presents a promising step towards articulated 3D shape creation at scale.
\section{Related Work}
\label{sec:related}

\paragraph{Part Segmentation}

In computer vision, most research on 2D or 3D part segmentation focuses on semantic ~\cite{yi2016scalable, mo2019partnet} or functional ~\cite{kim2014shape2pose, hu2018functionality, wang2019shape2motion, jiang2022opd} meanings of the parts. Jiang et al.~\cite{jiang2022opd} extended 2D semantic segmentation networks (e.g. Mask R-CNN) to predict the location of openable parts from a single-view image. Gelfand et al.~\cite{gelfand2004shape} defined a part segmentation as a kinematic surface composed of points that undergo the same type of motion. 
Other work has relied on a category-level canonical container or a fixed kinematic structure. For example,
follow-up work~\cite{li2020category, yan2019rpm} learned a fully supervised canonicalization to map instance observations of different position, orientation, scale and articulation into a canonical container. Xu et al.~\cite{xu2022unsupervised} estimated the part segmentation from a depth image in a known category of a fixed kinematic chain. 
In contrast, our approach requires neither semantic annotation nor category-level functional information or kinematic structure. Further, our work is not about learning semantic or functional part segmentation, but rather image-guided part segmentation. Our method finds the correspondence between an input image and a 3D shape under different articulation configurations. 

\begin{figure*}[t]
     \centering
     \includegraphics[width=0.9\textwidth]{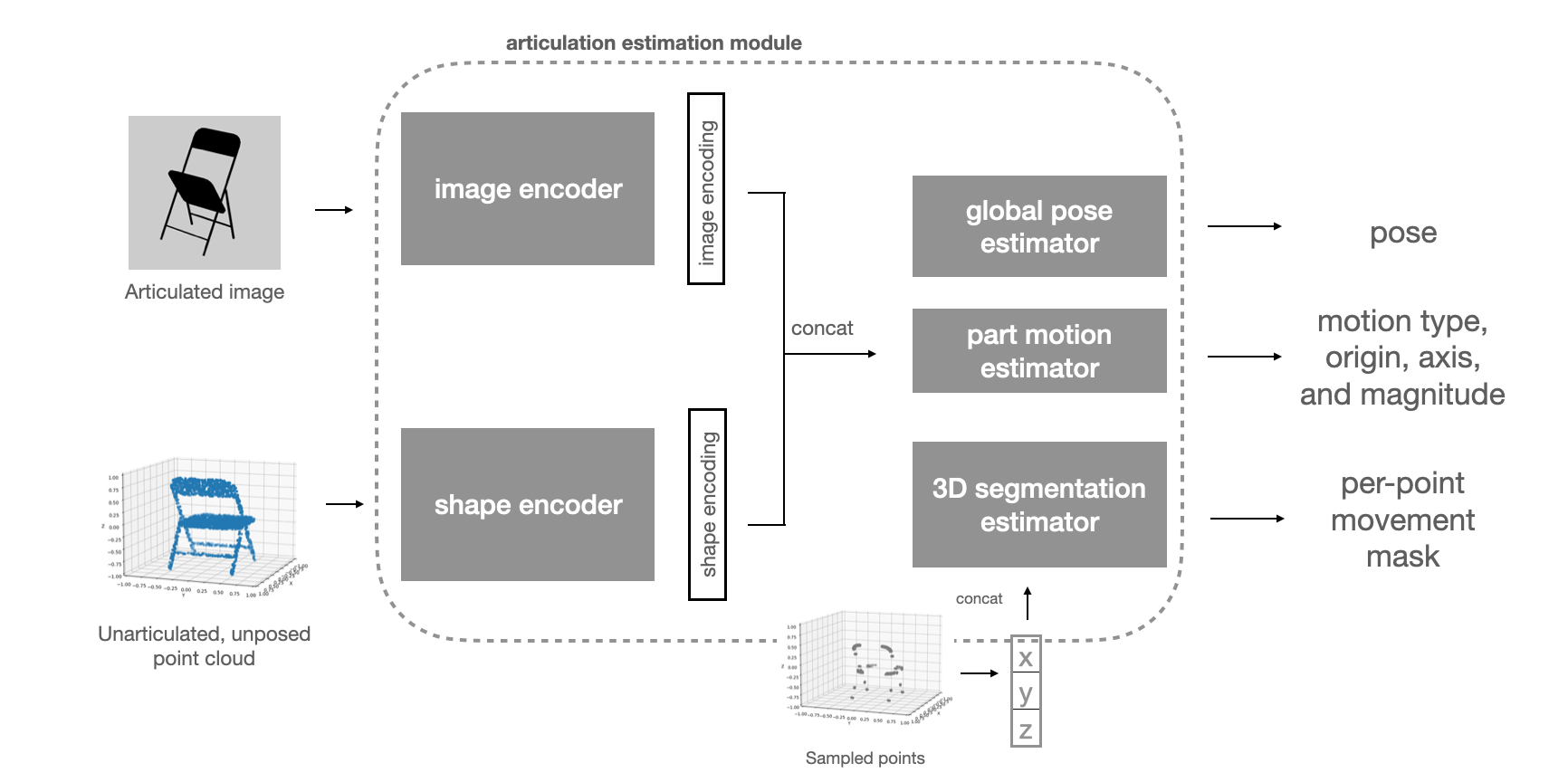}
   \vspace{-5pt}
   \caption{\textbf{Diagram of training architecture.} \catnet takes in an RGB image and 3D model and predicts the global pose of the object in the image, segmentation of the part that is articulated in the image, as well as part-specific motion parameters. Each motion attribute is predicted from a spearate head.}
\label{fig:architecture}
\end{figure*}

\vspace{-0.3cm}
\paragraph{Articulated Pose Estimation}
Articulated pose estimation from a 2D image is another focus of our work. 
Recently, Wang et al.~\cite{wang2019normalized} introduced the Normalized Object Coordinate Space (NOCS),
allowing category-level rigid pose estimation. Follow-ups~\cite{li2020category,  liu2022toward} have used this representation to perform articulated pose estimation. For example, Li et al.~\cite{li2020category} proposed a part-level canonical reference frame and performed part pose fitting using RANSAC. 
Liu et al.~\cite{liu2022toward} extended NOCS to Real-World Articulation NOCS which focuses on pose estimation for articulated objects with varied kinematic
structures in real-world settings. 
Kulkarni et al.,~\cite{kulkarni2020AA_CSM} introduced an approach for articulating a template mesh given a target image segmentation mask from a category-specific image collection, but requires hand-annotated part segmentations.
Other work~\cite{yan2019rpm, weng2021captra, yang2021lasr} relies on a series of frames displaying an object actively articulating. 
In contrast, our method does not require learning the canonical representation with a fixed kinematic structure for each category. Without the need of sequences of images of an object’s different articulation states as inputs, we require only an articulated single-view image and its unarticulated 3D model to perform articulated pose estimation. 

\vspace{-0.3cm}
\paragraph{Motion Estimation \& Transfer}
In 3D understanding, considerable work has focused on the problem of motion prediction. Methods largely differ in the types of input(s) used for motion estimation and transfer.
Initial work~\cite{gelfand2004shape, mitra2014structure, yan2019rpm} utilized only the geometry of an object to infer part segmentation and motion. 
Li et al.~\cite{li2016mobility} focused on predicting part mobility of an object from a sequence of scans displaying the the dynamic motion of articulated models.
More recently, motion estimation has been performed from depth images~\cite{abbatematteo2019learning, li2020category}. Mo et al.,~\cite{mo2021where2act} learned to make per-point motion predictions from a single RGB or depth image by interacting with articulated objects in simulation.
Other approaches use a sequence of images or video clips to infer 3D part movement~\cite{hu2017icon3, yi2018deep, jain2021screwnet}. Wei et al.,~\cite{wei2022self} trained an implicit representation of the geometry, appearance, and motion of an object in a self-supervised manner from image observations, however their approach is limited to certain classes of articulated objects.
Recently, Jiang et al.,~\cite{jiang2022opd} introduced a method for estimating 3D motion parameters from single image. Their approach 
aims to predict motion parameters for all ``openable" parts, rather than focusing on transferring arbitrary motion from one object to another.

Our work makes no assumptions on the input category and can both estimate and transfer 3D motion. We formulate the problem as learning the articulated segmentation of the input geometry, estimating the pose and predicting motion parameters of the moveable part (driven by the input image). Unlike many existing approaches, our work is both category-agnostic and structure agnostic.

\section{Method}

We propose a model for the \textit{single-view articulation transfer} task. Given an articulated RGB image and corresponding 3D model, our method infers the pose, part segmentation and motion parameters that can be used to deform the mesh to match the input image. 

\subsection{Architecture}
Our architecture is composed of 3 distinct branches: a pose prediction branch, a motion prediction branch, and a 3D part segmentation branch. Our model takes two inputs: an image $I \in \mathbb{R}^{h \times w \times 3}$ and point cloud shape $S \in \mathbb{R}^{n \times 3}$, where $n$ is the number of points in a point cloud sampled from the normalized input mesh. Image features, $\textbf{h} = f_\theta(I)$, are extracted with a ResNet-18 \cite{targ2016resnet}, and shape features, $\textbf{s} = g_\theta(S)$, are extracted with a PointNet-like \cite{qi2017pointnet} architecture, without input and feature transforms for pose-invariance. Both the image and shape encoder are trained from scratch. The shape and image features are concatenated and passed through a set of fully connected layers resulting in a joint image and shape embedding, $\textbf{z} \in \mathbb{R}^{b \times 512}$. Each of the three branches take $\textbf{z}$ as input. An overview of the full architecture can be found in Figure~\ref{fig:architecture}.
\vspace{-.3cm}

\begin{figure*}[t]
    \centering
    \includegraphics[width=0.9\linewidth]{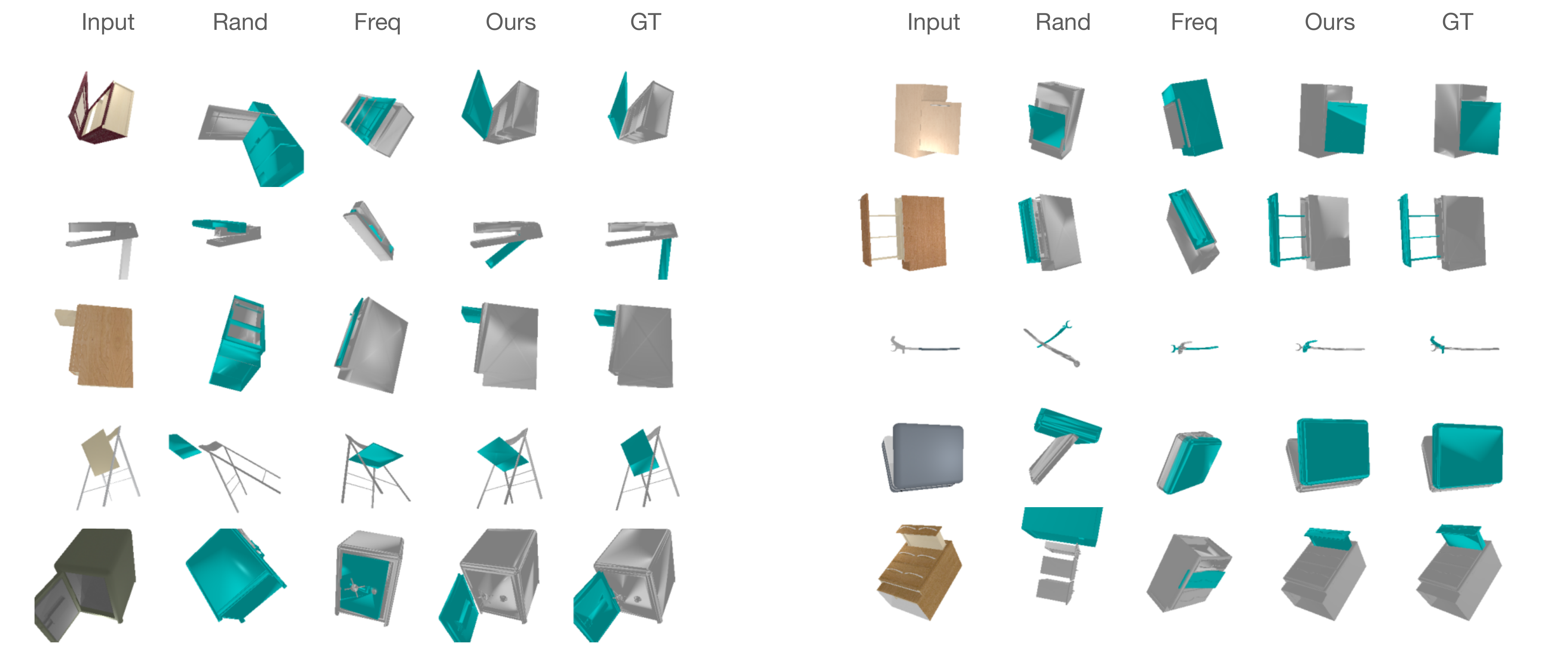}
    \caption{\textbf{Qualitative results on SAPIEN validation set.} Our method's predictions can be used to deform the input mesh to match the articulated image.}
    \vspace{-5pt}
    \label{fig:qual-sapien}
\end{figure*}
\paragraph{Pose Prediction}
The pose prediction branch of our architecture is inspired by \cite{xiao2019pose}, a method that also takes as input an RGB image (however unarticulated) and a 3D shape, and predicts the pose of the 3D model in the image. As in \cite{xiao2019pose} we predict pose in terms of azimuth, elevation, and in-plane rotation. The network first classifies the angle bin and the regresses an offset. Let $\hat{l}_{i,j}$ be the predicted bin for the $j$-th Euler angle (azimuth, elevation, in-plane rotation) of the $i$-th datapoint, and $\hat{\delta}_{i, j}$ be the predicted relative offset within the bins. For bin prediction, we have 24 azimuth classes, 12 elevation classes, and 24 in-plane rotation classes. We train the model with a cross-entropy loss ($\mathcal{L}_{CE}$) for bin classification and Huber loss ($\mathcal{L}_{huber}$) for regression offsets.
\begin{equation}
    \mathcal{L}_p = \sum_{i=1}^N \sum_{j \in \mathcal{E}} \mathcal{L}_{CE} (l_{i,j}, \hat{l}_{i,j}) + \mathcal{L}_{huber} (\hat{\delta}_{i, j}, \delta_{i, j}), 
\end{equation}
where $\mathcal{E} = \{\text{azimuth}, \text{elevation}, \text{in-plane}\}$ and $l_{i,j}$ and $\delta_{i,j}$ are the ground truth pose bins and offsets, accordingly.
\vspace{-.3cm}
\paragraph{Motion Estimation}
In this work we consider two distinct motion types, revolute (rotation) and prismatic (translation). We parameterize the revolute motion in terms of motion axis, origin, and magnitude, and prismatic in terms of motion axis and magnitude. Each motion attribute is predicted by a separate head, for origin, axis, magnitude, and class. Let $\mathcal{L}_{mc}$ be the classification loss on motion type, defined as:
\begin{equation}
    \mathcal{L}_{mc} = \sum_{i=1}^N \mathcal{L}_{CE} (m_i, \hat{m}_i),
\end{equation}
where $m_i$ is the ground-truth motion class represented as a one-hot vector and $\hat{m}_i$ is the corresponding predicted class.

Motion axis $\hat{a} \in \mathbb{R}^3$, origin $\hat{o} \in \mathbb{R}^3$, and magnitude $\hat{z} \in \mathbb{R}$ are regressed and optimized using Huber loss:
\begin{equation}
    \begin{split}
        \mathcal{L}_m = \sum_{i=1}^N \mathcal{L}_{huber} (a_i, \hat{a}_i) + \mathcal{L}_{huber} (o_i, \hat{o}_i) + \\ \mathcal{L}_{huber} (z_i, \hat{z}_i).
    \end{split}
\end{equation}
We predict motion parameters in the world coordinate space (i.e., object's canonical coordinate frame). Motion origin and magnitude for prismatic motion is given in terms of normalized object size, and magnitude for revolute motion is given in terms of radians. As motion origin is not relevant for prismatic joints, we only apply the motion origin loss, $\mathcal{L}_{huber} (o_i, \hat{o}_i)$, in the cases of predicted revolute motion.

\begin{table*}[t]
\begin{center}
{
    \setlength\tabcolsep{3pt}
    \ra{1.1} %
\begin{tabular}{llccccccccc}
  \toprule
  & & \multicolumn{7}{c}{Motion} & \multicolumn{2}{c}{3D} \\
  \cmidrule{4-8} \cmidrule{10-11}
    \multirow{8}{*}{\rotatebox[origin=c]{90}{I-split}}  \\
    \multirow{12.5}{*}{\rotatebox[origin=c]{90}{O-split}} 
  & & Pose ($\uparrow$) & Type ($\uparrow$) & Axis ($\downarrow$) & Origin ($\downarrow$) & Mag-R ($\downarrow$) & Mag-P ($\downarrow$) & Seg. ($\uparrow$) & Chamf. ($\downarrow$) & F1@0.1 ($\uparrow$) \\
  \midrule

  & RandMot & 12.6\% & 56.6\% & 78.8$^{\circ}$ & 1.09 & 62.0$^{\circ}$ & 0.746 & 54.5\% & 5.18 & 7.50 \\
  & FreqMot & 13.0\% & 69.0\% & 66.2$^{\circ}$ & 0.689 & 51.7$^{\circ}$ & 0.335 & 78.1\% & 3.51 & 9.09 \\

  & \catnet & 89.0\% & 99.2\% & 6.23$^{\circ}$ & 0.274 & 36.7$^{\circ}$ & 0.166 & 92.2\% & 1.05 & 42.5 \\
  \midrule
  & RandMot & 13.6\% & 58.6\% & 79.1$^{\circ}$ & 1.13 & 63.2$^{\circ}$ & 0.734 & 54.5\% & 4.79 & 7.57 \\
  & FreqMot & 13.2\% & 74.4\% & 67.3$^{\circ}$ & 0.792 & 54.7$^{\circ}$ & 0.333 & 76.6\% & 3.38 & 8.84 \\
  & \catnet & 74.0\% & 91.5\% & 58.7$^{\circ}$ & 0.605 & 51.9$^{\circ}$ & 0.230 & 81.8\% & 1.54 & 33.3  \\
  
  \bottomrule
\end{tabular}
}
\end{center}
\vspace*{-5mm}
\caption{\textbf{Quantitative performance of baselines and \catnet (ours) on SAPIEN validation split} Results for per-image (\textit{I-split}) and per-object (\textit{O-split}) dataset splits.
}
\vspace{-5pt}
\label{tab:quant-all-3d}
\end{table*}
\vspace{-.3cm}
\paragraph{3D Part Segmentation}
Our 3D part segmentation branch takes in the transformed global shape and image features, $\textbf{z}$, as well as a 3D point, $x \in \mathbb{R}^3$, and predicts whether or not that point (from the pointcloud) corresponds to the moveable part. Our architecture is inspired by DeepSDF \cite{park2019deepsdf} which also takes as input a set of features and a 3D point. After processing these inputs with 2 fully connected layers, the initial inputs are again concatenated with the intermediate features and further processed by 3 more fully connected layers. Let $\hat{p}$ be the predicted confidence of a certain point $x$'s movability - we define the 3D part segmentation loss as:

\begin{equation}
    \mathcal{L}_s = \sum_{i=1}^N \sum_{j=1}^M \mathcal{L}_{CE} (p_{i,j}, \hat{p}_{i, j}),
\end{equation}
where $i$ indexes over training datapoints and $j$ over the $M$ sampled 3D points from the input shape. For training we use $M = 1000$.
\vspace{-.4cm}
\paragraph{Full Training Objective}
Our full training loss is a weighted combination of losses applied to each of the three branches:
\begin{equation}
    \mathcal{L} = \lambda_p \mathcal{L}_{p} + \lambda_{mc} \mathcal{L}_{mc} + \lambda_{m} \mathcal{L}_m + \lambda_s L_{s}.
\end{equation}
The coefficients on the loss terms are $\lambda_{p} = 2$, $\lambda_{mc} = 1$, $\lambda_{m} = 8$, and $\lambda_{s} = 1$. We train the network using the Adam optimizer with a learning rate of 1e-3 and batch size of 128 for 50 epochs and decrease the learning rate to 1e-4 for the final 10 epochs.
\vspace{-.4cm}
\paragraph{Inference}
During inference, we randomly sample 100 points on each distinct part of the input model and compute the network's 3D part prediction for each point. We apply the network's predicted motion only to parts that have $>$50\% points considered ``moveable".
\section{Evaluation and Results}
\label{sec:results}

We train and evaluate our method using images and ground-truth data generated from the SAPIEN PartNet-Mobility~\cite{xiang2020sapien} dataset.

\begin{table*}[t]
\begin{center}
{
    \setlength\tabcolsep{3pt}
    \ra{1.1} %
\begin{tabular}{lccccccc}
  \toprule
  & & \multicolumn{5}{c}{Motion} & \\
  \cmidrule{3-7}
  & Pose ($\uparrow$) & Type ($\uparrow$) & Axis ($\downarrow$) & Origin ($\downarrow$) & Mag-R ($\downarrow$) & Mag-P ($\downarrow$) & Seg ($\uparrow$) \\
  \midrule

  Pose Only    & 69.6\% & -    & -              & -     & -                & -    & -   \\
  Motion Only  & -    & 92.5\% & 52.2$^{\circ}$ & 0.599 & 52.3$^{\circ}$ & 0.26 & -   \\
  Seg Only     & -    & -    & -              & -     & -                & -    & 78.9\% \\
  All          & 74.0 \% & 91.5\% & 58.7$^{\circ}$ & 0.605 & 51.9$^{\circ}$ & 0.23 & 81.8\%  \\
  \bottomrule
\end{tabular}
}
\end{center}
\vspace*{-5mm}
\caption{\textbf{Training for individual objectives vs multi-task learning.} We train three separate networks for pose, motion and segmentation only and compare their performance to the proposed multi-task objective. We find that optimizing all losses together yields the best results.
}
\vspace{-5pt}
\label{tab:branch-ablation}
\end{table*}

\begin{table}[h]
\begin{center}
{
    \setlength\tabcolsep{3pt}
    \ra{1.1} %
\begin{tabular}{lcccc}
  \toprule
  & \multicolumn{2}{c}{\quad \quad \quad \quad Motion} \\
  \cmidrule{2-4}
  & Type ($\uparrow$) & Axis ($\downarrow$) & Origin ($\downarrow$) & mAP ($\uparrow$) \\
  \midrule
  OPD-O~\cite{jiang2022opd}  & 94.0\%  & 72.5$^{\circ}$ & 0.912 & 41.0\% \\
  \catnet  & 98.9\%  & 50.5$^{\circ}$  & 0.757 & 49.2\% \\
  \bottomrule
\end{tabular}
}
\end{center}
\vspace*{-5mm}
\caption{\textbf{Comparison to OPD for 3D motion prediction and 2D segmentation}. We compare to OPD, a method that estimates motion parameters for \textit{all} movable parts given an RGB image.} For segmentation, we report mAP@IoU=0.5. 
\vspace{-5pt}
\label{tab:opd-baseline}
\end{table}

\subsection{Dataset}
The original SAPIEN PartNet-Mobility dataset consists of 2,346 CAD models from 46 categories with part segmentations and motion annotations. As we are interested in articulation transfer from a single-image, we focus on objects with large moveable parts such as the door or drawer of a cabinet, rather than the individual keys on a keyboard. To do so, we filter out CAD models that only contain moveable parts that consist of $<$5\% of the overall surface area of the object, as well as objects with missing or incomplete motion data. After this filtering process, 1,717 objects remain. 

From the remaining objects, we render 256 images per object using the SAPIEN Vulkan Renderer, each with a random pose, randomly sampled part, and randomly sampled motion parameters within the valid range. For simplicity, we only consider one moving part at a time. The pose distribution ranges from azimuth $\in [-90^\circ, 90^\circ]$, elevation $\in [-45^\circ, 45^\circ]$, and in-plane rotation $\in [-20^\circ, 20^\circ]$. Such ranges were chosen so that the articulated part on the 3D model is usually visible (most moveable parts are on the front of the object, rather than the back). For each image, we also randomly sample the camera's field-of-view $\in [20^\circ, 60^\circ]$. In total we render approximately 280K images for training, and 10K images for evaluation.

For each image we also write out the rest-state mesh, the ground truth pose of the object, 3D part segmentation (of the articulated part from the rest of the object), and motion parameters. The motion parameters consist of a motion type $\in \{revolute, prismatic\}$, axis, origin, and magnitude. In the case of revolute motion, the magnitude corresponds to a rotation amount, whereas it corresponds to translation distance for prismatic motion. From the rest-state mesh we sample a point cloud that is the input to our network. For training, we resize images to 224 $\times$ 224 and sample a pointcloud of 2500 points for each mesh. Pointclouds are further normalized to fit in a unit box.

\subsection{Baselines}
We consider three baseline approaches to single-view articulation transfer: selecting a random motion and part (RandMot), selecting the most frequent motion parameters (FreqMot), and OPD-O~\cite{jiang2022opd}, the current state-of-the-art approach for 3D motion prediction. RandMot and FreqMot are inspired by similar baselines used by OPD~\cite{jiang2022opd}, however modified to predict additional outputs such as object pose and motion magnitude. RandMot gives a lower bound on performance whereas FreqMot is a relatively strong heuristic-based baseline approach.
\vspace{-0.3cm}
\paragraph{RandMot}
For the random motion (RandMot) baseline, we sample a random set of ground truth labels from the training dataset, including random pose, motion axis, motion origin, type, and magnitude. For the predicted part segmentation, we randomly sample a valid part from the set of parts contained in the input mesh.
\vspace{-0.3cm}
\paragraph{FreqMot}
In contrast to the RandMot baseline, the frequent motion (FreqMot) baseline method predicts the most frequent motion parameters from the train dataset. We still randomly sample a global pose, but sample the predicted part segmentation based on statistics from the training dataset. We compute the average size of the most frequently sampled part in the training set and select the part that is most similar in size at inference time.
For the remaining categorical variables, we sample the most frequent value from the training set. For real-valued parameters such as motion magnitude, motion axis, and motion origin, we first cluster the values from the training set and sample the mean value of the most frequent cluster.
\paragraph{Openable Part Detection~\cite{jiang2022opd}}
While designed for a different problem formulation, Openable Part Detection (OPD) is the most applicable to our articulation transfer task in terms of model output. OPD takes a single RGB image and predicts 2D segmentation masks and 3D motion parameters for all parts identified as ``openable". 
While their method cannot be directly used to animate a 3D model, we can still compare to them in terms of 3D motion prediction and 2D part segmentation. Specifically, we compare our method to OPD-O, the best-performing variant of OPD in terms of real image performance. OPD-O also predicts motion parameters in the world coordinate system by using an additional branch to predict extrinsic parameters.

\subsection{Metrics}
We consider evaluation metrics for pose, motion, and segmentation performance, as well as 3D reconstruction metrics on the final deformed mesh. Our \textit{Pose} metric is the percentage of pose estimations with rotation error less than 30$^{\circ}$. \textit{Motion-Type} refers to the 2-way classification accuracy of revolute or prismatic motion. \textit{Motion-Axis} is the average angular error (in degrees) between the normalized predicted and ground-truth motion axes, and \textit{Motion-Origin} is the average $L_1$ distance between the predicted and ground-truth motion origin. \textit{Mag-R} and \textit{Mag-P} refer to the revolute and prismatic magnitude errors, respectively, measured in degrees for the case of revolute motion and $L_1$ distance for prismatic motion. Because motion origin is not applicable for prismatic motion, we only evaluate motion origin for predicted revolute parts. \textit{Seg.} refers to the 3D segmentation accuracy, measured across 1000 sampled points. Finally for 3D reconstruction metrics, we report Chamfer distance and F1@0.1.

\subsection{Experimental Results}
\paragraph{Performance on Validation Set}
We report the performance of our model trained on the SAPIEN dataset in Table~\ref{tab:quant-all-3d}. As no neural network-based approach currently exists for the full single-view articulation transfer task, we compare against our two heuristic-based baselines: RandMot and FreqMot. We consider two train/validation splits of varying difficult, \textit{I-split} (per-image: validation set contains unseen images) and \textit{O-split} (per-object: validation set contains unseen images of novel objects). 

Compared to the RandMot and FreqMot baselines, we find that our method achieves the best performance across all metrics. Understandably, the validation performance is better on the I-split than the O-split. Visualizations of each method's predictions can be found in Figure~\ref{fig:qual-sapien}. 
\vspace{-0.3cm}
\paragraph{Comparison to OPD} %
We compare our approach to the recently introduced OPD method~\cite{jiang2022opd}, which 
represents the state-of-the-art in terms of 3D motion estimation. OPD predicts 2D segmentation masks, motion type, and 3D motion parameters (axis, origin) for \textit{all} moveable parts (based on priors in the training dataset) given a single image, whereas \catnet takes a 3D model and a driving input image, and predicts 3D segmentation, motion type, and 3D motion parameters (axis, origin, \textit{and} magnitude). To compare the two approaches on a somewhat level playing field, we evaluate the performance of \catnet and OPD on the
motion parameter predictions that the methods share in common. Because OPD makes multiple predictions for all moveable parts in an image but in our problem setting we are only interested in the part that has actually moved in the image, we take only the OPD predictions that have $\geq$ 0.5 IOU with the ground-truth segmented part. Further, as OPD is limited to predicting only \textit{openable} part motion, we only consider datapoints that come from the subset of the 11 categories that OPD was trained on. This reduces the size of our valiation set from 10,000 images to 1,039 images. We utilize the \catnet model trained on the more difficult per-object training split (O-split).

\begin{figure}[t]
    \centering
    \includegraphics[width=1.0\linewidth]{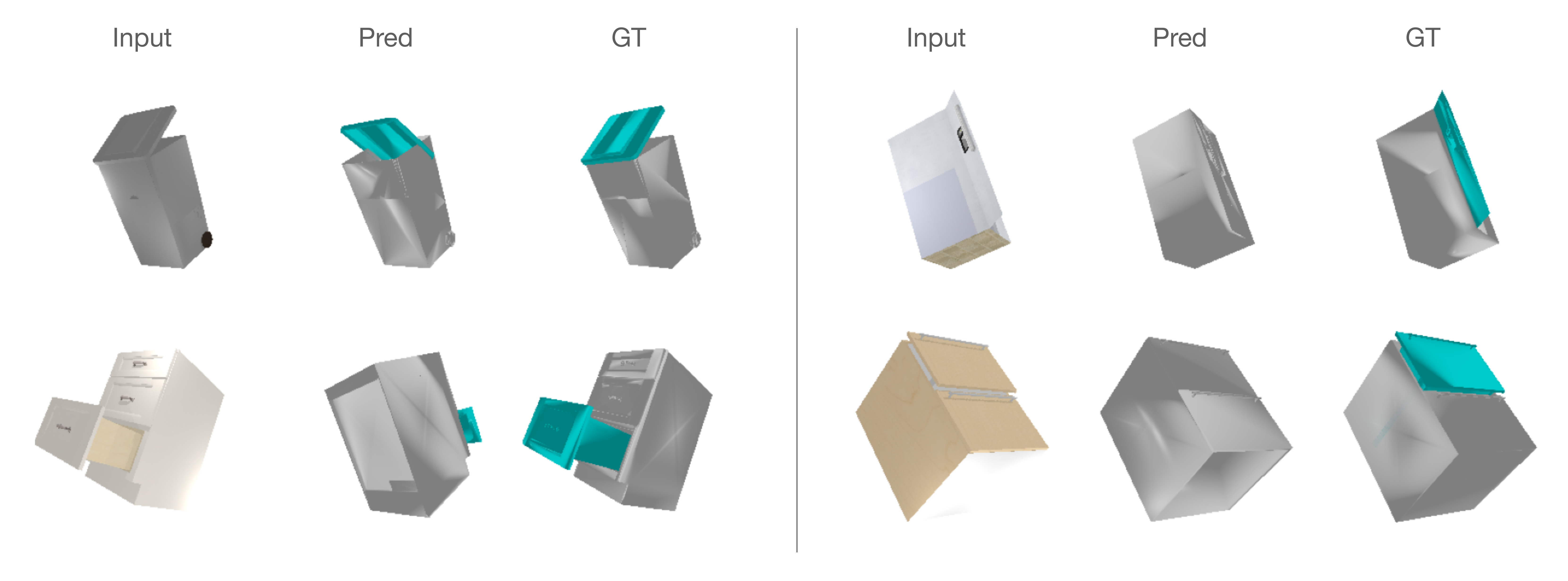}
    \caption{\textbf{Typical failure cases in articulation transfer}. Pose prediction failures (180 degrees off, left column), and failure to predict a segmented part when motion magnitude in image is small (right column).}
    \label{fig:failure-cases}
\end{figure}

Results can be found in Table~\ref{tab:opd-baseline}. We also compare to OPD in terms of 2D segmentation ability, by projecting the \catnet-predicted 3D part segmentation to 2D and measuring mAP at IoU = 0.5. In general we find that \catnet outperforms OPD in terms of motion type, origin, and axis prediction, as well as 2D segmentation ability. Intuitively, learning 3D features from the input model in \catnet should support better prediction of 3D motion parameters. A common failure case that we observe with OPD is predicting a motion axis that is 180 degrees rotated from the correct axis. \\ \\
\noindent \textbf{Multi-Task vs Single-Branch Learning}
We study the contribution of each of the loss terms corresponding to each branch of our architecture by training three variants of our model: pose prediction only, motion prediction only, and 3D segmentation prediction only. The results can be found in Table~\ref{tab:branch-ablation}. We find that the multi-task objective with all three loss terms leads to the best performance across most metrics, specifically for pose and 3D segmentation estimation. The performance gains for motion estimation are less pronounced. Intuitively, accurate pose prediction should aid the alignment between the image and shape inputs, which should improve part segmentation ability. Such an effect may explain the boost in performance by doing multi-task training. \\ \\
\noindent \textbf{Failure Cases}
The most common failure cases of our method include pose prediction failure (e.g., predicting a pose that is 180-rotated from the ground-truth pose) as well as failure to segment any part of the object when the motion magnitude is small. Visualizations of these failure cases can be found in Figure~\ref{fig:failure-cases}.

\begin{figure}[t]
    \centering
    \includegraphics[width=1.0\linewidth]{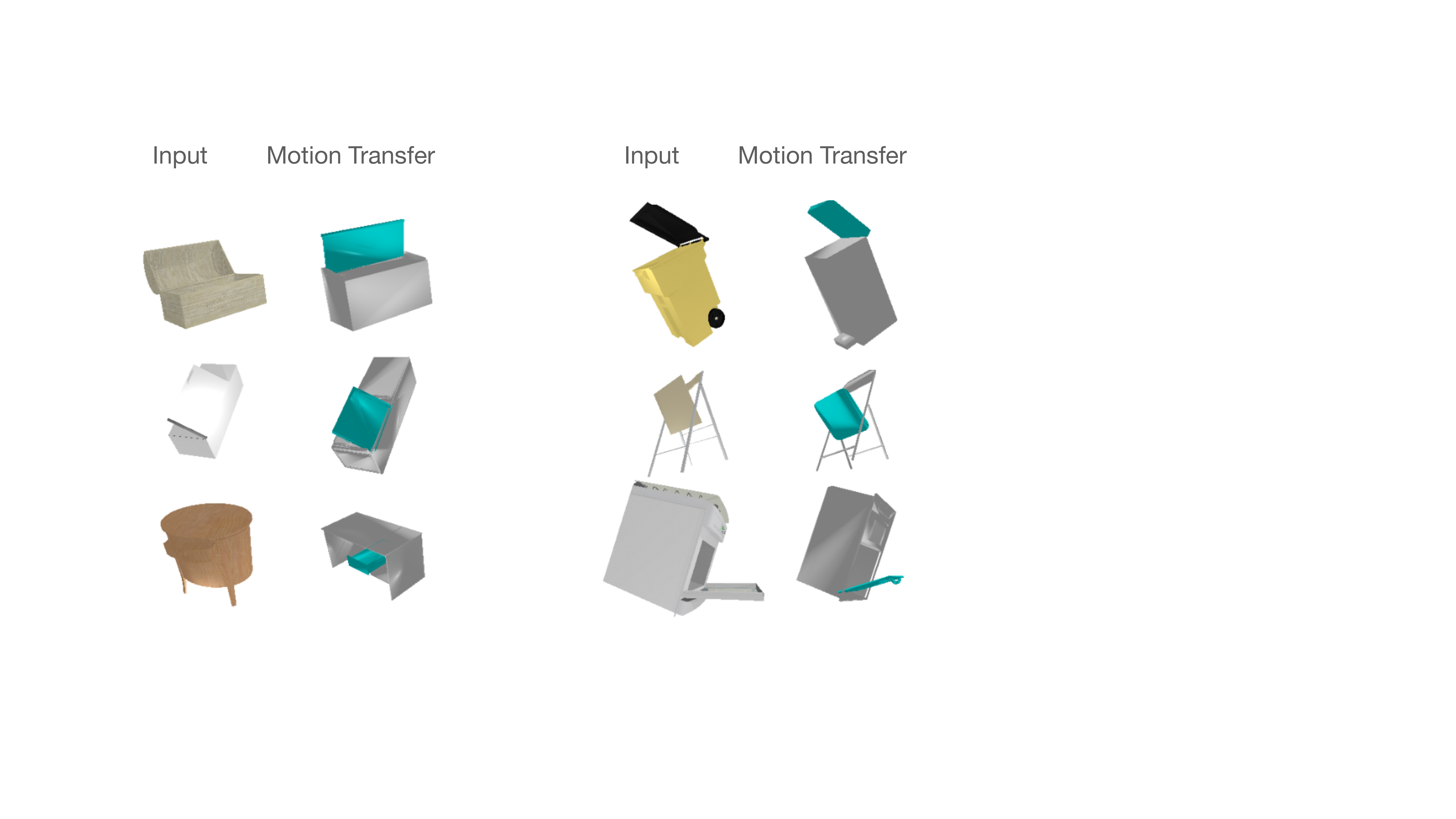}
    \caption{\textbf{Motion analogies.} By presenting a functionally similar but geometrically distinct 3D model at test time, we can transfer motion across analogous 3D parts.}
    \vspace{-5pt}
    \label{fig:shape-transfer}
\end{figure}

\section{Applications}

We present three exciting applications of our model: generalization to functionally similar objects (i.e., motion analogies), 3D model animation, and transfer to real images and meshes in ABO.

\subsection{Motion Analogies}
During training, our method takes as input a 3D model that exactly matches the geometry of the object pictured in the 2D image. However, at test time we can swap the 3D input with a functionally similar 3D model that does not exactly match the input image to perform a \textit{motion analogy}. To achieve this, we swap the input model with a different model from the same category as the original input 3D model, and run our method with no additional modifications. The articulated outputs predicted by our method can be seen in Figure~\ref{fig:shape-transfer}. We note that in this setting our model has to not only segment and predict motion for the corresponding articulated part on the swapped shape, it also has to accurately predict the pose in this setting that was never seen during training. In general we find that our method can segment the functionally similar part and produce motion parameters that lead to a plausible deformation of the non-matching input model, even when the input model is drastically different from the articulated image (see Figure~\ref{fig:shape-transfer} left column, third row).

\subsection{Animation: Interpolation and Extrapolation}
\begin{figure}[t]
    \centering
    \includegraphics[width=0.9\linewidth]{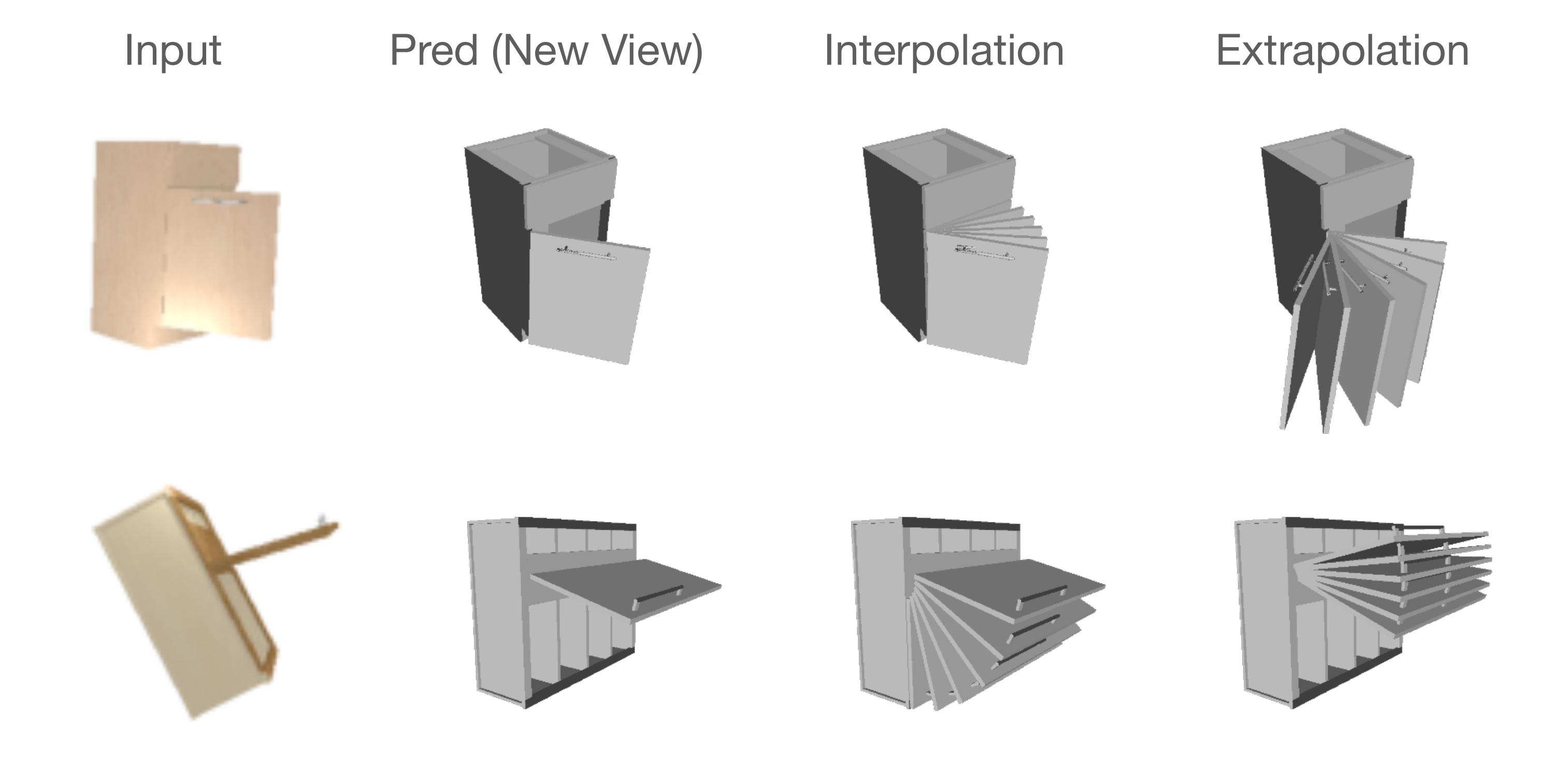}
    \caption{\textbf{Interpolation and extrapolation in 3D from a single input image.} Our motion parameterization can easily be used to interpolate between the rest-state and articulated object pose, or even extrapolate beyond the motion that is seen in the input image.}
    \vspace{-5pt}
    \label{fig:animation}
\end{figure}

A convenience of our motion parameterization is that animation of the input 3D model can be achieved simply by varying the predicted motion magnitude. As a result, the predicted segmented part will move along or around the predicted axis. We can interpolate between the rest-state magnitude (0) and predicted magnitude to simulate the motion of a part opening and closing, or even consider magnitude settings beyond the predicted value to get motion extrapolation. See Figure~\ref{fig:animation} for an animation example.

\subsection{Transfer to Real Objects \& Images}
A natural source of real-world articulated objects, including rest-state mesh and image pairs, comes from the ABO dataset \cite{collins2022abo}. The dataset contains catalog images of Amazon products, as well as a subset of 3D models corresponding to the object in the images. The catalog images range from stock-like photos to close-up detail shots, and also sometimes contain an articulated picture of the object. We gathered a subset of approximately 50 catalog images that show articulated objects (with a single moving part) for which there also exist corresponding 3D models as a real-world use-case of the single-view articulation transfer setup.

Unlike the training objects in the SAPIEN dataset that have annotated part segmentations, meshes from the ABO dataset are unsegmented. Further, articulated image instances in ABO have no ground-truth in terms of pose, part segmentation, and motion annotations. To achieve a candidate part segmentation we leverage the fact that the meshes in ABO happen to be designed by artists, part by part, and thus different parts of the object tend to be represented by mesh pieces that are disconnected from the rest of the object. As a result, running connected components on the ABO meshes tends to give us an over-segmentation of the object's functionally relevant parts. We treat these connected components as pseudo-part segmentations and can then run \catnet on this data with no modifications. 

\begin{figure}[t]
    \centering
    \includegraphics[width=0.9\linewidth]{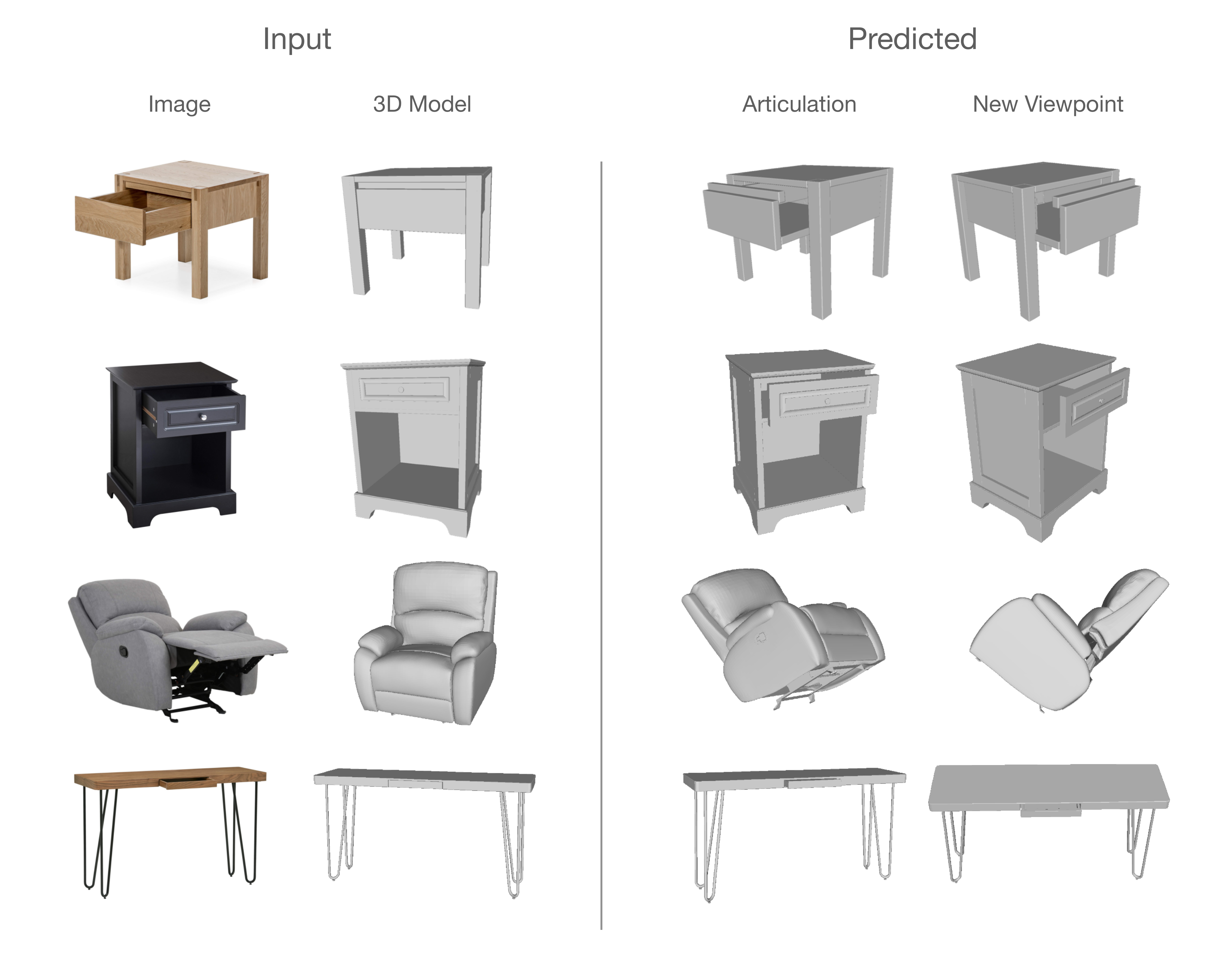}
    \caption{\textbf{Transfer results to real articulated objects.} We run our model on instances of ABO objects with no finetuning and are able to make reasonable predictions despite the domain gap.}
    \vspace{-5pt}
    \label{fig:abo-transfer}
\end{figure}

Qualitative results showing our model applied, without any finetuning, to real images from the ABO dataset can be seen in Figure~\ref{fig:abo-transfer}. Note that the meshes in ABO were designed to simply show the product in its rest-state, rather than with articulation in mind (like in SAPIEN). This is shown by the fact that drawers are simply external panels rather than a full drawer part enclosed by the rest of the object. Further, our model was trained only with inputs that show a single articulated part, whereas realistic catalog images typically show multiple articulated parts at once. We include a challenging instance of this in Figure~\ref{fig:abo-transfer} (row 3), where the input image contains a recliner chair, which is a category \textit{not seen during training}, with multiple complex motions, including a reclined headrest and raised footrest. As our method is category-agnostic, it can handle inputs like this at test-time.

\section{Conclusion and Future Work}
\label{sec:future}
We proposed a method to transfer the articulation from a single-view image to a 3D model in such a way that the model can be deformed to match the input image. To our knowledge, we are the first to solve this task in the setting of household objects from arbitrary categories.
In this work we only consider articulated exemplar images with a single moving part displayed. Of course, articulated images can commonly show multiple moving parts. Extending our method to predict multiple motion types and part segmentations is a practical next step and can be achieved by combining Mask RCNN-style detections (as used in OPD) with our method. Further, our method requires a 3D model as input which can limit its practical use in say, an embodied perception system navigating the world. Recent advances in CAD model retrieval from a single image~\cite{kuo2020mask2cad, kuo2021patch2cad, gumeli2022roca} can be used to retrieve a similar 3D CAD model, and our results on functionally similar shape swapping suggest that our approach will be able to handle this case. Thus combining \catnet with CAD retrieval methods could be a promising path forward to allowing our method to run on single image inputs only. Finally, we consider only rigid motion (further, just prismatic and revolute motion) in this work, but considering how this approach can be extended to more complicated motion types as well as non-rigid deformation is an interesting avenue for future work.

\newpage

{\small
\bibliographystyle{ieee_fullname}
\bibliography{egbib}
}

\clearpage
\setcounter{section}{0}
\renewcommand\thesection{\Alph{section}}
\section{Additional Architecture Details}
Here, we outline the architecture used for each predictive branch in \catnet. Let $FC(in, out)$ refer to a fully-connected layer with $in$ input and $out$ output units. $BN$ refers to a Batch Normalization layer and $ReLU$ is a rectified linear unit nonlinearity. We process the input image with a ResNet-18 and point cloud with a PointNet-style architecture to get intermediate features \textbf{h} $\in \mathbb{R}^{b \times 256}$ and \textbf{g} $\in \mathbb{R}^{b \times 256}$. We concatenate these features to form $\textbf{z} \in \mathbb{R}^{b \times 512}$, a joint image-shape embedding. These global features are further processed by the following three branches:

\paragraph{Pose Prediction}
The pose prediction branch first further compresses the global features (via $f_{compress}(\textbf{z})$, defined below) and then uses six separate linear readout layers to predict azimuth, elevation and in-plane rotation bins and offsets from the further compressed global features.

\noindent $f_{compress}(\cdot) = FC(512,800) - BN - ReLU - FC(800, 400) - BN - FC(400, 200) - BN - ReLU$

\paragraph{Motion Estimation}
Motion predictions are made directly from global features $\textbf{z}$, with a separate linear layer for each motion attribute. Motion class is predicted by passing $\textbf{z}$ through $FC(512, 2)$, motion axis by $FC(512, 3)$, motion origin by $FC(512, 3)$ and motion magnitude by $FC(512, 1)$.

\paragraph{3D Part Segmentation}
The segmentation branch takes in global features, as well as 3D coordinates, $x$. The architecture is as following:

\noindent $g_{compress}(\cdot) = FC(512 + 3, 256) - BN - ReLU - FC(256, 256) - BN - ReLU$ \\
$g_{seg}(\cdot) = FC(512 + 3 + 256, 256) - BN - ReLU - FC(256, 256) - BN - ReLU - FC(256, 2)$ \\

\noindent The final per-point prediction, $\hat{p}$, is made by:

$$ \hat{p} = g_{seg} ( [ g_{compress} ([\textbf{z}, x]]), \textbf{z}, x ] ), $$

where $[\cdot]$ represents a concatenation operation across the feature dimension.

\section{Articulated vs. Unarticulated Pose \\ Estimation}
Our pose prediction branch is largely inspired by the architecture of PoseFromShape~\cite{xiao2019pose}, which introduces a model for pose estimation that also takes as input an RGB image and template 3D shape. However, this work is focused on pose estimation when the input image is unarticulated and generally matches the configuration of the rest-state mesh. We wondered about the relative difficulty of predicting the global pose for images of articulated objects vs. images of unarticulated objects (as in the PoseFromShape setting). To investigate this, we rendered additional images of the SAPIEN dataset with unarticulated input images. We trained PoseFromShape for pose estimation on synthetic SAPIEN data in the case of both articulated, and unarticulated input images. We measured a validation Acc30 of 88.6\% in the case of unarticulated images (the standard PoseFromShape setting), compared to an Acc30 of 82.9\% for articulated images. Given the reasonably small performance drop ($\sim$6\%) going from articulated to unarticulated pose estimation, we concluded that performing articulated pose estimation is not that much more difficult of a task. For example, the PoseFromShape network may learn to pay attention to more robust cues such as parallel lines in the silhouette of the object (rather than positioning of individual object parts) in order to perform global pose estimation, which are not typically affected much by articulation of single object parts.

\begin{figure}[t]
    \centering
    \includegraphics[width=1.0\linewidth]{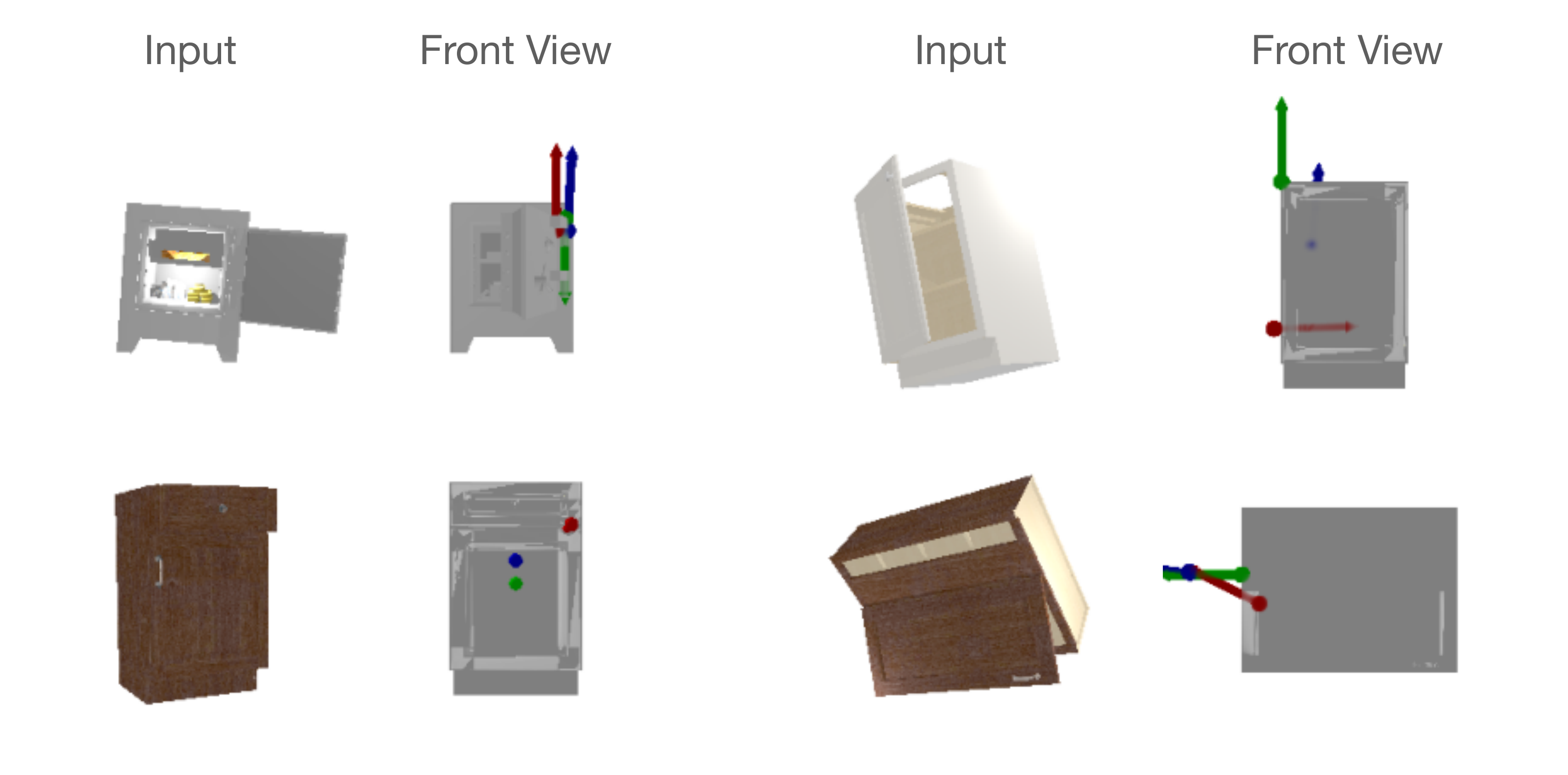}
    \caption{\textbf{Motion axis visualization for our network vs. OPD.} Input image as well as frontal view of 3D model shown with colored arrows corresponding to motion predictions. \catnet predicted motion axis and origin in {\color{Blue} \textbf{blue}}, OPD in {\color{BrickRed} \textbf{red}}, and ground-truth in {\color{OliveGreen} \textbf{green}}.
    }
    \vspace{-5pt}
    \label{fig:supp-opd}
\end{figure}

\section{Additional Qualitative Results}
We show additional qualitative results for \catnet trained on the SAPIEN dataset. We visualize 3D motion axes predicted by \catnet (trained on the \textit{O-split}) and OPD, show connected component segmentations of ABO meshes, and include additional qualitative results for \catnet trained on the SAPIEN \textit{I-split}.

\paragraph{Visualization of Motion Axes}
Using the prediction motion axis and origin values, we can visualize them as arrows on the 3D shape. Figure~\ref{fig:supp-opd} shows examples of such motion predictions for \catnet and OPD. We also plot the ground truth motion axis and origin. Meshes are visualized in their default, front-facing pose with colored arrows that begin at the predicted origin, and point in the direction of the predicted axis.

\begin{figure}[t]
\vspace{-12cm}
    \includegraphics[width=1.0\linewidth]{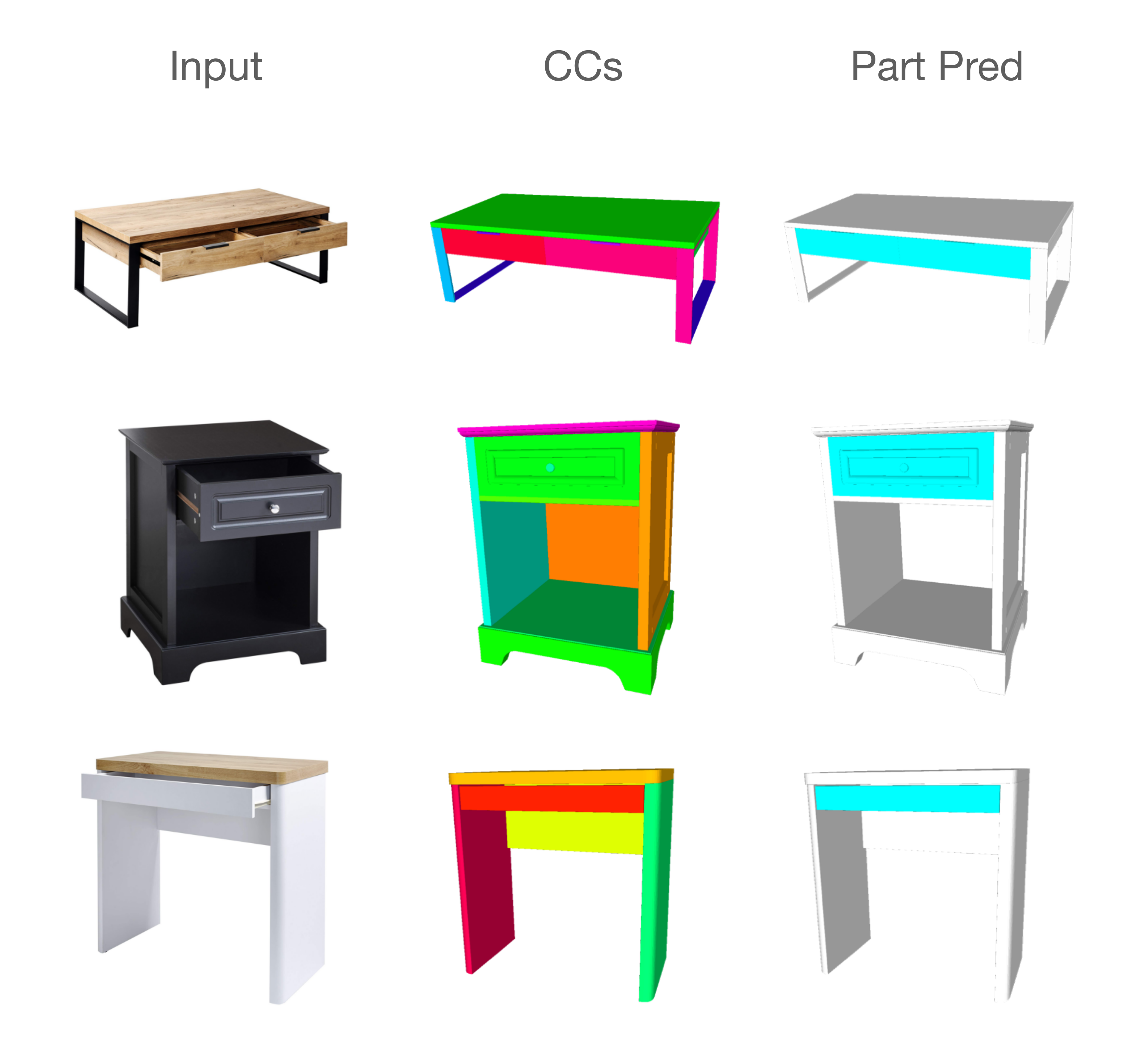}
    \caption{\textbf{Visualization of connected components in ABO.} Middle column shows 3D mesh corresponding to input image with each connected components visualized as a different color. Our method samples points from each candidate part (i.e., connected component) and returns a part prediction corresponding to the component that moved in the input.}
    \vspace{-5pt}
    \label{fig:supp-abo-cc}
\end{figure}

\paragraph{ABO Connected Components}
Unlike synthetic SAPIEN models, ABO meshes do not contain ground-truth part segmentations. We found that a good proxy for these is a connected components-based segmentation of the mesh. In general, connected components will give an over-segmentation of the model (e.g. each piece of wire in a wire shelf would be its own component), however our method is still able to select the relevant 3D part conditioned on the articulated input image. We visualize ABO meshes colored by their connected components, as well as our model's predicted part segmentation in Figure~\ref{fig:supp-abo-cc}.

\begin{figure*}[t]
    \centering
    \begin{center}
    \includegraphics[width=0.8\linewidth]{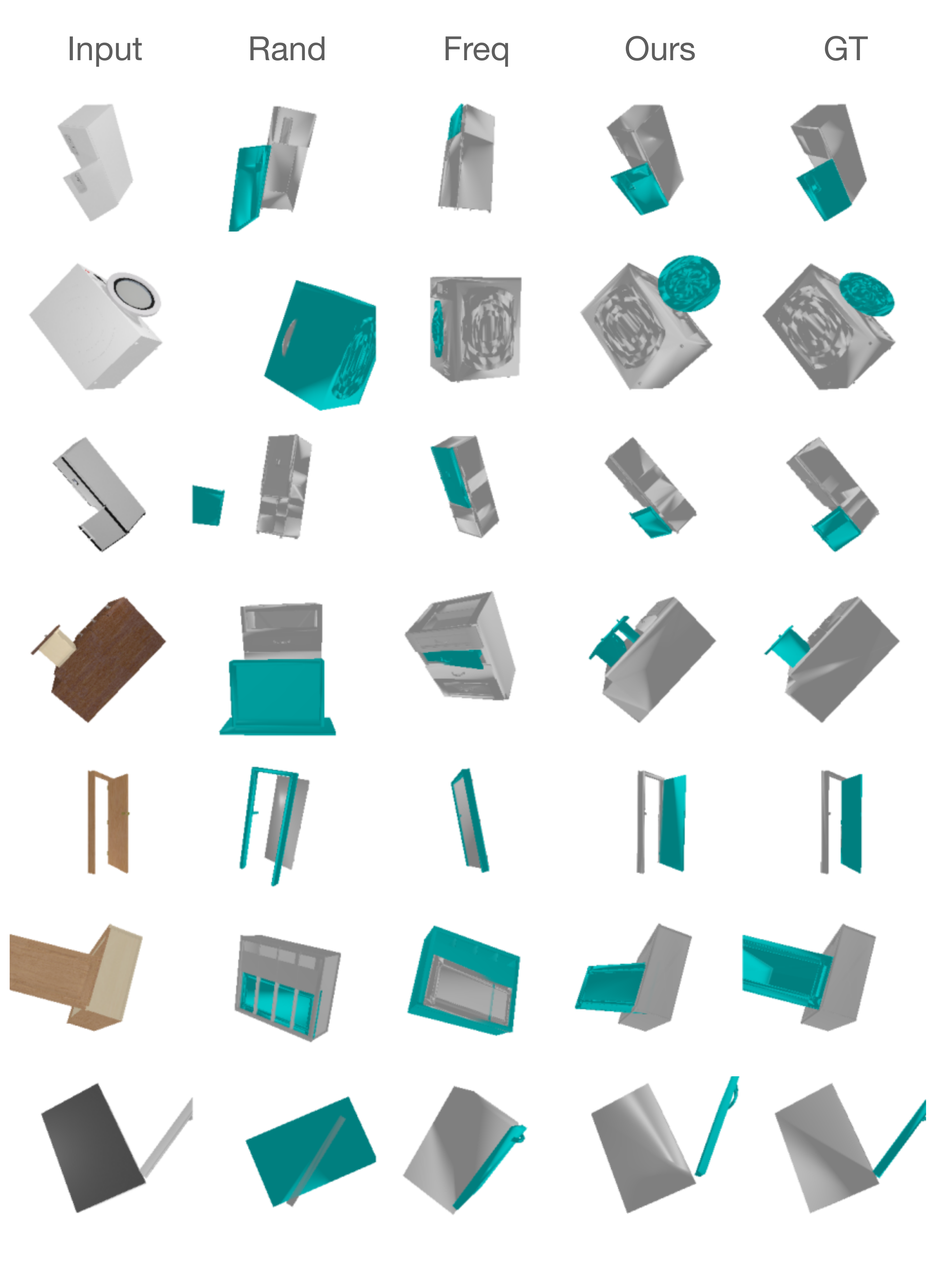}
    \end{center}
    \caption{\textbf{Additional qualitative results on SAPIEN validation set.} Our method compared to the RandMot and FreqMot baselines.}
    \vspace{-5pt}
    \label{fig:supp-qual}
\end{figure*}

\paragraph{SAPIEN Validation Set}
We provide additional qualitative results of \catnet trained on the SAPIEN \textit{I-split} in Figure~\ref{fig:supp-qual}. This visualization shows renderings of the resulting articulated 3D meshes, articulated by our model's predictions. We compare to the RandMot and FreqMot baseline.

\newpage

\end{document}